\pdfoutput=1
\documentclass[11pt]{article}

\usepackage{EMNLP2023}

\usepackage{times}
\usepackage{latexsym}

\usepackage[T1]{fontenc}

\usepackage[utf8]{inputenc}

\usepackage{microtype}
\usepackage{algorithm}
\usepackage{algorithmic}
\usepackage{amsmath}
\usepackage{hyperref} 
\usepackage{stfloats}
\usepackage{booktabs}
\usepackage{subfigure}
\usepackage{cleveref}
\usepackage{xcolor}
\usepackage{multirow}
\usepackage{graphicx}
\usepackage{enumitem}
\usepackage{hyperref}

\usepackage{inconsolata}

\title{Code Search Debiasing:\\Improve Search Results beyond Overall Ranking Performance}

\author{
    Sheng Zhang$^{1}$, ~~Hui Li$^{1}$\thanks{$\quad$Corresponding Author.}, ~~Yanlin Wang$^{2}$, ~~Zhao Wei$^{3}$, ~~Yong Xu$^{3}$, ~~Juhong Wang$^{3}$\\ \bf{Rongrong Ji}$^{1}$ \\
    $^1$Key Laboratory of Multimedia Trusted Perception and Efficient Computing\\Ministry of Education of China, Xiamen University \\
    $^2$School of Software Engineering, Sun Yat-sen University, $^3$Tencent \\
    \texttt{sheng@stu.xmu.edu.cn}, \texttt{wangylin36@mail.sysu.edu.cn} \\
    \texttt{\{zachwei,rogerxu,julietwang\}@tencent.com}, \texttt{\{hui,rrji\}@xmu.edu.cn} \\ 
}

\begin{document}
\maketitle


\begin{abstract}
	Code search engine is an essential tool in software development. Many code search methods have sprung up, focusing on the overall ranking performance of code search. In this paper, we study code search from another perspective by analyzing the bias of code search models. Biased code search engines provide poor user experience, even though they show promising overall performance. Due to different development conventions (e.g., prefer long queries or abbreviations), some programmers will find the engine useful, while others may find it hard to get desirable search results. To mitigate biases, we develop a general debiasing framework that employs reranking to calibrate search results. It can be easily plugged into existing engines and handle new code search biases discovered in the future. Experiments show that our framework can effectively reduce biases. Meanwhile, the overall ranking performance of code search gets improved after debiasing. Our implementation is available at: \url{https://github.com/KDEGroup/CodeSearchDebiasing}.
\end{abstract}

\section{Introduction}

Software development is a repetitive task as programmers usually reuse or get inspiration from existing implementations.
Studies show programmers spent 19\% of their programming time on searching source code~\citep{BrandtGLDK09}.  
Therefore, code search, which refers to the retrieval of relevant code snippets from a codebase according to programmer's intent that has been expressed as a query~\citep{LiuXLGYG22}, has become increasing important~\citep{abs-2204-02765}.

Although much effort has been devoted to improving code search, existing works mostly emphasize the ranking performance of code search w.r.t. metrics like Mean Reciprocal Rank (MRR) and Hit Ratio@K (HR@K)~\cite{LiuXLGYG22,abs-2204-02765}.
In this paper, we study code search from another perspective.
We find that state-of-the-art code search methods prevalently have discriminatory behaviors (i.e., different performance) toward queries or code snippets with certain properties (e.g., length). 
The observation shows, even though the overall ranking performance is good, programmers may still be dissatisfied with search results when their input queries or desired code snippets fall into those categories that code search models cannot handle well.
We name our observation as \emph{Code Search Bias}, inspired by the AI bias that attracts great attention recently~\citep{MehrabiMSLG21}.  
Code search bias hurts user experience.
Due to different development conventions (e.g., prefer long queries or abbreviations), users (programmers) of
code search engines with biases will have different user experience, i.e., some users will find the engine useful,
while others may find it hard to get desirable search results.

Note that most studies of bias in NLP focus on societal bias~\cite{BlodgettBDW20}.
For example, the gender bias of NLP algorithms may pose the danger of giving preference to male applicants in automatic resume filtering systems~\citep{SunGTHEZMBCW19}. 
However, in applications like search engines~\cite{OvaisiAZVZ20} and recommender systems~\cite{LinLXL21,XvLLSYC22}, some biases without societal factors are widely studied as they make the system biased toward certain search results and harm the performance. 
For instance, position bias exists in learning-to-rank systems where top search results are more likely to be clicked even if they are not the most relevant results~\cite{AgarwalZWLNJ19,XvLLSYC22}.
But it does not mean any discriminatory behaviors toward certain groups of people.
Similarly, code search bias does not involve societal factors.

Considering that our observation has revealed the widespread code search bias in existing models, we aim at designing a general debiasing framework that can be easily plugged into existing code search engines.
In the context of code search bias, debiasing indicates removing the correlations between code search quality and certain properties of queries and code snippets.
Our proposed debiasing framework adopts the idea of reranking to calibrate search results. 
It helps state-of-the-art code search models overcome code search bias and their overall performance can be improved at the meantime. 
In summary, our contributions are:
\begin{enumerate}[leftmargin=13pt,topsep=1pt,itemsep=0.2pt] 

\item To our best knowledge, we are the first to study code search bias. We reveal the widespread existence of seven code search biases.

\item To mitigate code search bias, we propose a general debiasing framework using reranking. It can be easily plugged into existing engines.

\item Extensive experiments show that our debiasing framework not only helps alleviate code search bias but also improves the overall ranking performance of state-of-the-art code search models.

\end{enumerate} 


\section{Related Work}

\vspace{3pt}
\noindent\textbf{Code Search.} 
Early code search methods adopt traditional information retrieval methods to estimate 
the relevance between the query and a code snippet~\citep{LvZLWZZ15,BajracharyaOL10}.
Recent works adopt deep neural networks to embed query and code into vectors.
Then, the code search task is performed by measuring the similarity between vectors.
Along this direction, various deep learning based methods have been proposed, 
including but not limited to 
recurrent neural network (RNN) based approaches~\citep{DeepCS}, 
convolutional neural network (CNN) based approaches~\citep{CQIL}, 
graph neural network (GNN) based approaches~\citep{WanSSXZ0Y19}
and pre-training approaches~\citep{CodeBERT, GraphCodeBERT, GuoLDW0022}.

\vspace{5pt}
\noindent\textbf{Bias and Debias.}
Many AI systems exhibit certain biases that bring unfairness and degrade the performance~\citep{MehrabiMSLG21}. 
Various debiasing methods have been proposed and they can be roughly divided into three types: 
\begin{enumerate}[leftmargin=12pt,topsep=1pt,itemsep=0.2pt] 
	\item \textbf{Pre-processing methods} remove biases in training data. \citet{CalmonWVRV17} design a framework for discrimination-preventing pre-processing to enhance data with multi goals. \citet{BiswasR21} analyze bias prompts in data preprocessing pipelines and identify data transformers that can mitigate the pipeline bias. 

	\item \textbf{In-processing methods} mitigate biases in the model training step. \citet{DGarimellaAKYNCS21} propose a debiasing method that requires pre-training on an extra small corpus with bias mitigation objectives for mitigating social biases in language models. \citet{LinLXL21} propose a debiasing framework with three strategies that be used as regularizers in the training objective of review-based recommender systems. 

	\item \textbf{Post-processing methods} handle biases after model training. \citet{PetersenMSY21} translate debiasing into a graph smoothing problem and propose a post-processing coordinate descent algorithm. \citet{KimGZ19} design Multiaccuracy Boost, which uses an auditor to identify subpopulation biases and further uses it for debiasing in the post-processing. 
\end{enumerate}

Although many debiasing methods exist, they cannot be directly used for code search biases.
Our method belongs to the post-processing category and it is tailored for removing code search biases.


\section{Analysis of Code Search Biases}
\label{sec:bias}

\subsection{Analysis Settings}

\vspace{3pt}
\noindent\textbf{Data:} We use CoSQA dataset\footnote{\href{https://github.com/microsoft/CodeXGLUE/tree/main/Text-Code/NL-code-search-WebQuery}{https://github.com/microsoft/CodeXGLUE/tree/main/Text-Code/NL-code-search-WebQuery}}~\citep{CoCLR} with 20,604 query-code pairs. Each query is written in English while each code snippet is a Python code snippet. 
The data is annotated by at least 3 human annotators. 
We randomly split the dataset by 70\%/30\% for training and test. 
We adopt byte-pair encoding tokenization, a standard tokenization method used in preprocessing code search data, to tokenize queries and code snippets.
As queries are typically short, stop words in queries are not removed.
Note that there are other public code search datasets, e.g., CodeSearchNet dataset~\citep{abs-1909-09436}, DeepCS dataset~\citep{DeepCS}, and CodeXGLUE dataset~\citep{LuGRHSBCDJTLZSZ21}.
We choose CoSQA dataset as it includes real code search queries, while other datasets use code documents (e.g., the first sentence in the function comments) to mimic queries.
Using CoSQA helps us better discover biases in a real code search scenario.

\vspace{3pt}
\noindent\textbf{Code Search Models:} We select six representative code search approaches in the literature for our bias analysis, including DeepCS\footnote{\href{https://github.com/guxd/deep-code-search}{https://github.com/guxd/deep-code-search}}~\citep{DeepCS}, CQIL\footnote{\href{https://github.com/flyboss/CQIL}{https://github.com/flyboss/CQIL}}~\citep{CQIL}, CodeBERT\footnote{\href{https://github.com/microsoft/CodeBERT}{https://github.com/microsoft/CodeBERT} \label{footnote:codebert}}~\citep{CodeBERT}, CoCLR\footnote{\href{https://github.com/Jun-jie-Huang/CoCLR}{https://github.com/Jun-jie-Huang/CoCLR}}~\citep{CoCLR}, GraphCodeBERT$\textsuperscript{\ref{footnote:codebert}}$~\citep{GraphCodeBERT} and UniXcoder$\textsuperscript{\ref{footnote:codebert}}$~\citep{GuoLDW0022}. They are all under the MIT license, allowing us to adopt them in this study. We have observed similar biases in all the six methods. Due to space limitation, we only show analysis results of CQIL, CodeBert and GraphCodeBERT, and other methods are reported in our debiasing experiments in Sec.~\ref{sec:exp}. We follow authors' descriptions to set hyper-parameters whenever possible in order to tune the performance of each method towards its best.

\vspace{3pt}
\noindent\textbf{Evaluation Metrics:} We use Mean Reciprocal Rank (MRR), the most widely used measure for code search, to illustrate our bias analysis. It is defined as $\text{MRR} = \frac{1}{\vert Q \vert} \sum_{i=1}^{\vert Q \vert} \frac{1}{rank_i}$,
where $\vert Q \vert$ is the number of queries and $rank_i$ indicates the rank of the ground-truth code snippet w.r.t. the $i$-th query.
We also adopt another prevalent metric Hit Ratio@K (HR@K, the percentage of ground-truth code snippets that are in the top-$K$ ranking lists from code search models) and results are discussed in Sec.~\ref{sec:exp}.
Note that most current code search studies assume that there exists only one good result for each query and public code search datasets are designed this way.
Hence, the popular ranking metric Normalized Discounted Cumulative Gain (NDCG) will be consistent with MRR.
Our reported results are averaged over several runs.

\subsection{Analysis Results}

\begin{figure*}[t]
	\centering
	\includegraphics[width=1\linewidth]{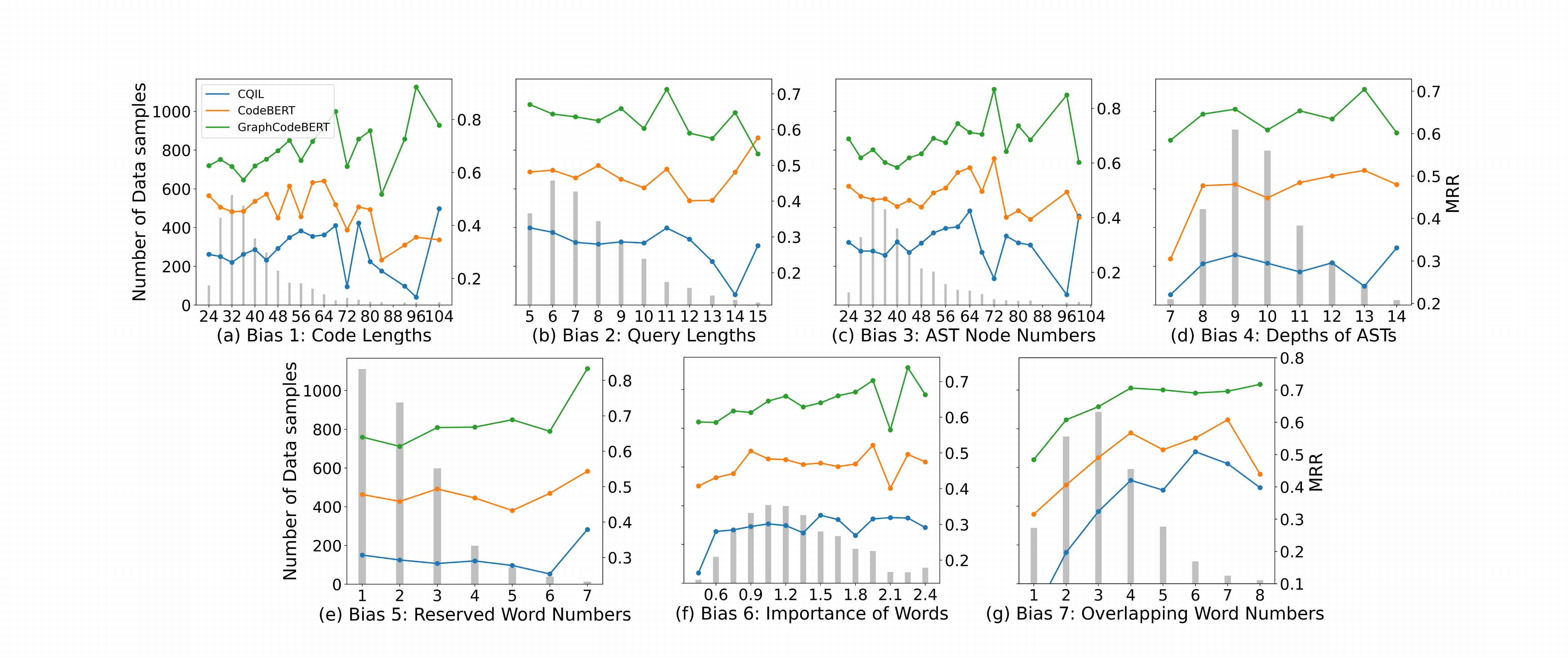}
	\vspace{-20pt}
	\caption{Examples of code search biases.}
	\vspace{-15pt}
	\label{fig:Code search Bias}
\end{figure*}

Based on the characteristics of code search and the data involved in the search process, we have found and verified seven code search biases. 
A general motivation to consider these seven factors is that they are commonly adopted as parameters in the experiments of existing papers as they affect the results of code-related tasks~\cite{abs-2208-11271,WanZYXY0Y18,McBurneyM16}. 
The performance of CQIL, CodeBERT and GraphCodeBERT w.r.t. the seven biases are presented in Fig.~\ref{fig:Code search Bias}.
We first group queries (in the test set) or ground-truth code snippets in intervals with equal lengths w.r.t. certain statistics. 
Then, we investigate whether code search models show different behaviors towards different intervals.
The x-axis illustrates the intervals.
To better visualize the result of bias analysis, data in Fig.~\ref{fig:Code search Bias} (a) and (c) is grouped in an interval with a length of 4, data in Fig.~\ref{fig:Code search Bias} (f) is grouped in an interval with a length of 0.15, and data in other subfigures is grouped in an interval with a length of 1.
The left y-axis denotes the number of queries or ground-truth code snippets in each interval while the right y-axis shows the average MRR score for data in each interval.
We provide our analysis as follows:

\vspace{3pt}
\textbf{Bias 1 w.r.t. Lengths of Ground-Truth Code}
\vspace{1pt}

Length bias (i.e., model makes decisions based on or affected by the length of texts) has been verified in various information retrieval and natural language processing tasks such as textual matching~\citep{JiangLLMLY22} and machine translation~\citep{MurrayC18}.
This inspires us to investigate the effect of the length of ground-truth code snippets on code search models.

Fig.~\ref{fig:Code search Bias} (a) shows the performance of three models w.r.t. code lengths. 
From Fig.~\ref{fig:Code search Bias} (a), we can see that lengths of most code snippets are between 20 and 50.
Furthermore, we can observe that: 
(1) In general, the longer the ground-truth code snippet is, the better the MRR score is. There are some sharp drops in MRR when code length gets much longer. The reason may be the number of ground-truth code snippets in intervals with longer lengths (e.g., $>70$) is quite small and a few hard cases affect the average performance in those intervals.
(2) Code search models show a clear bias towards intervals with longer lengths of ground-truth code snippets, i.e., longer ground-truth code snippets are more easily to match.
For instance, the MRR scores of GraphCodeBERT are 0.57 and 0.83 for the interval with average code length 36 and the interval with average code length 68, respectively. 
Intuitively, longer ground-truth code snippets provide more semantic information, making it more easy to be modeled and matched. 
From a software engineering perspective, long code snippets are more distinctive than short ones: it is more likely for two short code snippets to be similar, making it hard to distinguish the correct one from other candidates. 

\vspace{3pt}
\textbf{Bias 2 w.r.t. Lengths of Queries}
\vspace{1pt}

Similar to Bias 1, we have identified the bias w.r.t. lengths of input queries.
As shown in Fig.~\ref{fig:Code search Bias} (b),  
as query length increases, MRR decreases, indicating that longer queries have worse search results. 

\vspace{3pt}
\textbf{Bias 3 w.r.t. Numbers of AST Nodes}
\vspace{1pt}

One major difference between natural languages (NLs) and programming languages (PLs) is that PLs have strict syntax rules that are enforced by language grammars.
Abstract Syntax Tree (AST), used in compilers, represents the abstract syntactic structure of the source code.
Each node of ASTs denotes a construct or symbol occurring in the source code.
Compared to plain source code, ASTs are abstract and some details (e.g., punctuation and delimiters) are not included.
ASTs are used in various code-related tasks like code summarization~\cite{LinOZCLW21}, code completion~\citep{WangL21a}, issue-commit link recovery~\cite{abs-2308-10759} and refactoring~\cite{LiuWWXWLJ23} for capturing syntactic information.

Considering the importance of ASTs for modeling PL syntax, we investigate the influence of ASTs on code search models.
Usually, longer code snippets correspond to deep ASTs. 
However, some complex yet short code snippets such as list parsing in Python may also have deep ASTs. 
Hence, Bias 3 is not equivalent to Bias 1.
Fig.~\ref{fig:Code search Bias} (c) demonstrates the impacts of AST node numbers on the performance of code search models. 
We can observe the bias: code search models show diverse performance towards different intervals. 
For example, the MRR scores of GraphCodeBERT are 0.6 and 0.87 for the interval with average AST node number 40 and the interval with average AST node number 72, respectively.
The performance gap is significant in code search.

\vspace{3pt}
\textbf{Bias 4 w.r.t. Depths of ASTs}
\vspace{1pt}

Similar to Bias 3, we further identify the bias w.r.t. AST depths which also depict the complexity of ASTs.
Note a deep AST may not have many AST nodes. Hence, Bias 3 and Bias 4 are different.
Fig.~\ref{fig:Code search Bias} (d) shows the impact of AST depths. 
In Fig.~\ref{fig:Code search Bias} (d), code snippets are grouped by the depth of their ASTs and the interval length is 1. 
We can observe the existence of bias: code search models have diverse performance towards different intervals containing ASTs with different depths.

\vspace{3pt}
\textbf{Bias 5 w.r.t. Numbers of Reserved Words}
\vspace{1pt}
 
If we do not consider identifiers and constants, the vocabulary of code tokens containing reserved words of a PL is small.
We investigate the impact of reserved words on the behaviors of code search models.
Specially, we consider Python reserved words \texttt{if}, \texttt{for}, \texttt{while}, \texttt{with}, \texttt{try} and \texttt{except}. 
They are related to control structures and demonstrate the programming logic of designing a function.
Fig.~\ref{fig:Code search Bias} (e) demonstrates the performance towards ground-truth code snippets containing different numbers of reserved keywords.
We can see the existence of a bias: performance of code search models varies when the number of code keywords changes.
We can observe that the considerable growth of the MRR score when the number of keywords in ground-truth code snippets increases.
One possible reason is that logic-related reserved words in ground-truth code snippets help code search models better capture the logic of the code. Therefore, it is easier for code search models to match the ground-truth code snippet and the user intent that manifests in the queries when code contains more logic-related reserved words.

\vspace{3pt}
\textbf{Bias 6 w.r.t. Importance of Words}
\vspace{1pt}

Queries are typically concise, containing only a few words.  
For each query, we calculate the max TF-IDF values for the words contained in the query to estimate how important words contained in a query are. 
We have also calculated the average and the minimum TF-IDF values and similar results can be observed.
TF-IDF helps avoid amplifying the importance of words that appear more frequently in general (e.g., the word ``an'' in a query ``sort an array'').
When calculating TF-IDF, we treat each query in CoSQA as a document.
Results are presented in Figs.~\ref{fig:Code search Bias} (f), and we can observe the existence of a bias, i.e., code search models show different performance for queries containing words with varying importance.
Intuitively, the important words (e.g., ``sort'') contained in a query help code search models better understand user intent and match the ground-truth code snippet.

\vspace{3pt}
\textbf{Bias 7 w.r.t. Numbers of Overlapping Words}
\vspace{1pt}
 
Early code search methods rely on the overlapping words of queries and code snippets to estimate query-code relevance scores.
However, overlapping words received less attention in deep learning based code search models~\citep{ZhuSLXZ20}.
We investigate the influence of overlaps on the behaviors of the three code search models which all leverage deep learning.
Fig.~\ref{fig:Code search Bias} (g) illustrates the performance on test query-code pairs that have different numbers of overlapping words.
From the figure, we can observe a bias: models produce better MRR towards query-code pairs with more overlapping words.
In other words, deep learning-based code search models also capture overlapping words and treat them as a strong signal of a matching result, confirming the standard hypothesis that overlapping words affect code search.
\vspace{3pt}
In summary, we have identified seven distinct biases, meaning that code search models show different performance when facing input queries or ground-truth code snippets with different characteristics.
In practice, code search biases result in the inconsistence of user experience: depending on the characteristics of queries and/or ground-truth code snippets, the quality of search results varies.


\section{Mitigate Code Search Biases}

In this section, we illustrate our debiasing framework shown in Fig.~\ref{fig:debiasing}.
Our goal is to design a \emph{general} framework: (1) it can be easily plugged into existing code search models without much additional effort, and (2) it can handle new code search biases that are not discovered at the moment. 

We opt to adopt \emph{reranking}, a post-processing method, to calibrate code search results.
The idea is to rerank the ranking results provided by code search models.
Even though code search biases are prevalent in many cases as we have seen in Fig.~\ref{fig:Code search Bias}, many code search models show promising overall performance (i.e., high MRR or HR@K).
Therefore, for biased cases,  
the ground-truth code snippets are not too far away from the top of search results.
Otherwise, the overall MRR scores will be quite low according to its definition.
Similarly, we believe that any new code search biases also meet the above condition (i.e., biases exist but overall search performance is high).
For biased cases, a successful reranking method can help ground-truth code snippets emerge on top.
Post-processing search results also avoid modifying existing code search models.
This way, the designed debiasing framework is orthogonal to a specific code search method and it can be easily used as a reinforcement.

Next, we first demonstrate how our framework mitigates one bias. 
Then, the way that our framework mitigates multiple biases is presented.

\subsection{Mitigate A Single Bias via Reranking}
\label{sec:single_bias}
 
Our idea is to use the prior knowledge of biased search from the training data to determinate whether a similar search in the test set will face a bias issue and require reranking.
The detailed steps of mitigating a single bias via a single reranker are:
\begin{enumerate}[leftmargin=12pt,topsep=1pt,itemsep=0.2pt]
\item Firstly, we embed all queries in the training set into vectors using a pre-trained CodeBERT model. For a test query (i.e., the current search), after it is embedded by the CodeBERT model, we retrieve its top-$M$ most similar queries in the training set based on cosine similarity between vectors. These retrieved queries and their corresponding ground-truth code snippets in the training set will provide some hints on whether the current search may face a certain bias.

\begin{figure}[t]
    \centering
    \includegraphics[width=0.98\columnwidth]{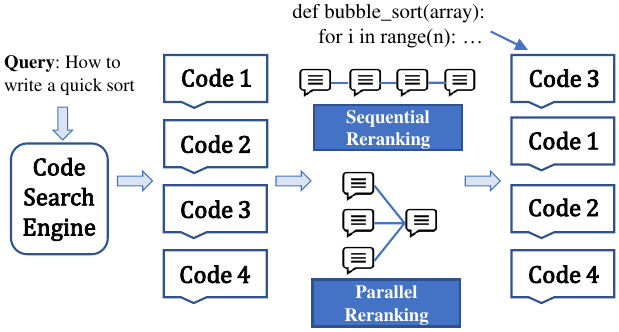}
    \vspace{-10pt}
    \caption{Overview of the debiasing framework.}
    \vspace{-15pt}
    \label{fig:debiasing}
\end{figure}

\item Then, we identify intervals in training data where code search models show \emph{very high} performance. It is likely that search results are not severely biased within these intervals. Otherwise the MRR scores for these intervals should be low by its definition. For such intervals, it is unnecessary to rerank for debiasing. 
We sort the search cases in training set by their MRR scores and retrieve cases with top $N\%$ maximum MRR scores.
We adopt k-means to cluster the retrieved training search cases into $S$ clusters.
Then, the maximum and minimum MRR scores in each cluster are used as the boundaries of the cluster.

\item For a test search $t$, if its top-$M$ most similar training query-code pairs have an average MRR score that falls in the range of any cluster, then it is likely that code search models provide reasonable relevance prediction scores for the candidate code snippets contained in these query-code pairs and our method will not rerank these candidate code snippets.
For other candidate code snippets, reranking is required.

\item For a candidate code snippet $c$ that requires reranking, the reranking score is calculated as:
\vspace{-10pt}
\begin{equation}
    \label{rerank eq}
    R = \text{Score}_{c}^{\text{original}} + P(T_{e} < T_{m}), 
\end{equation}
where $\text{Score}_{c}^{\text{original}}$ denotes the original ranking score of $c$, $T_e$ represents the MRR value of the code search model on a training query-code pair, 
$T_m$ represents the overall MRR value of the code search model on the training data,
and $P(T_{e} < T_{m})$ indicates the percentage of training query-code pairs that the code search model shows a lower MRR score than its overall MRR score over all the training pairs.

\item For the test search $t$, our method will use reranking scores $R$ instead of $\text{Score}^{\text{original}}$ as relevance scores for all candidate code snippets that are identified to require reranking in Step 3. Then, the ranking list is reranked according to new relevance scores.

\end{enumerate}
We discuss the impact of the choices of $M$, $N$ and $S$ in Analysis 5 of Sec.~\ref{sec:exp}.

\subsection{Mitigate Multiple Biases}

To mitigate multiple code biases together, we adopt two simple yet effective strategies to assemble re-rankers for different code search biases:
\begin{enumerate}[leftmargin=12pt,topsep=1pt,itemsep=0.2pt] 
\item \textbf{Sequential Reranking}: Adopt each reranker sequentially. The relevance scores from a previous reranker will be used as the base relevance scores (i.e., $\text{Score}^{\text{original}}$) in the next reranker.
\item \textbf{Parallel Reranking}: Adopt each reranker parallel and use the average of the relevance scores from all rerankers between a candidate code snippet and the current search as the prediction.
\end{enumerate}

\begin{table}[t]
\centering
\caption{Comparisons of two reranking methods. ``S'' and ``P'' indicate sequential reranking and parallel reranking, respectively. $R_1$ and $R_2$ are reranking scores from reranker 1 and reranker 2, respectively.}
\label{tab:rerank_example}
\vspace{-10pt}
\scalebox{0.52}{
\renewcommand{\arraystretch}{1.5}
\begin{tabular}{|c|c|l|l|l|}
\hline
\textbf{Method}    & \textbf{Code} & \multicolumn{1}{c|}{\textbf{Reranker 1}} & \multicolumn{1}{c|}{\textbf{Reranker 2}}     & \multicolumn{1}{c|}{\textbf{Relevance Score}}     \\ \hline
\multirow{2}{*}{S} & c1            & $\text{Score}^{\text{original}}_{c1}+R_1$ & $\text{Score}^{\text{original}}_{c1}+R_1+R_2$ & $\text{Score}^{\text{original}}_{c1}+R_1+R_2$     \\ \cline{2-5} 
                   & c2            & $\text{Score}^{\text{original}}_{c2}$     & $\text{Score}^{\text{original}}_{c2}$         & $\text{Score}^{\text{original}}_{c2}$             \\ \hline
\multirow{2}{*}{P} & c1            & $\text{Score}^{\text{original}}_{c1}+R_1$ & $\text{Score}^{\text{original}}_{c1}+R_2$     & $\text{Score}^{\text{original}}_{c1}+(R_1+R_2)/2$ \\ \cline{2-5} 
                   & c2            & $\text{Score}^{\text{original}}_{c2}$     & $\text{Score}^{\text{original}}_{c2}$         & $\text{Score}^{\text{original}}_{c2}$             \\ \hline
\end{tabular}
\renewcommand{\arraystretch}{1}
}
\vspace{-15pt}
\end{table}

\vspace{5pt}
Tab.~\ref{tab:rerank_example} provides examples to illustrate relevance scores between a query and two candidate code snippets $c_1$ and $c_2$.
From final relevance scores of the code snippet $c_1$, we can see that sequential reranking emphasizes the adjustment of reranking as it aggregates reranking terms from different rerankers.
Differently, parallel reranking averages reranking terms from different rerankers, avoiding a sharp reranking.
If none of the rerankers adjust the relevance score, then the final relevance scores are the same for both methods, as shown in the case of the code snippet $c_2$.
Empirically, different ordering shows only slight performance difference, as we will show in Analysis 4 of Sec.~\ref{sec:exp}.

\begin{figure*}[!th]
    \centering
    \includegraphics[width=1\linewidth]{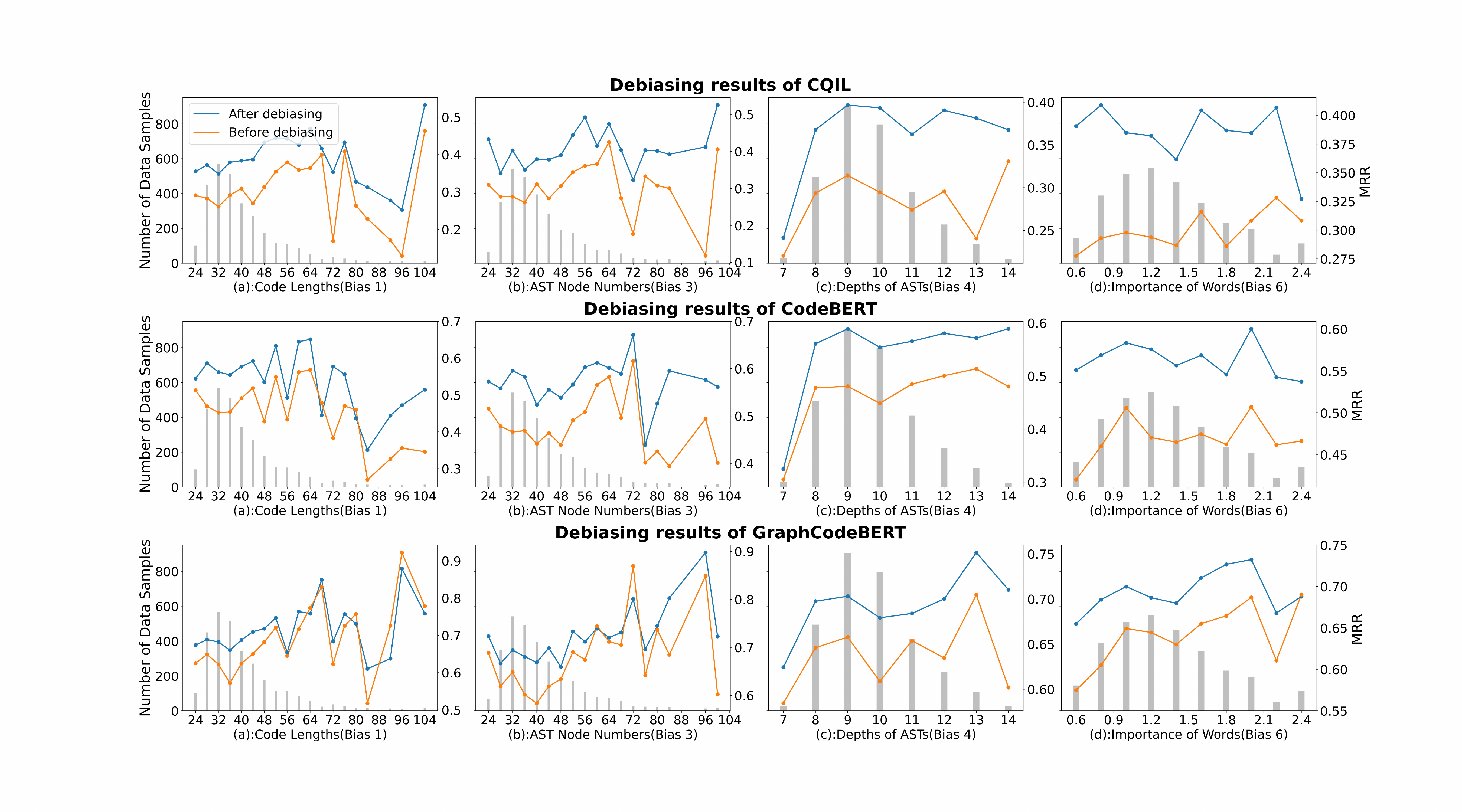}
    \vspace{-20pt}
    \caption{Mitigate biases using sequential reranking.}
    \vspace{-5pt}
    \label{fig:debiasing_seq}
\end{figure*}

\begin{figure*}[!th]
    \centering
    \includegraphics[width=1\linewidth]{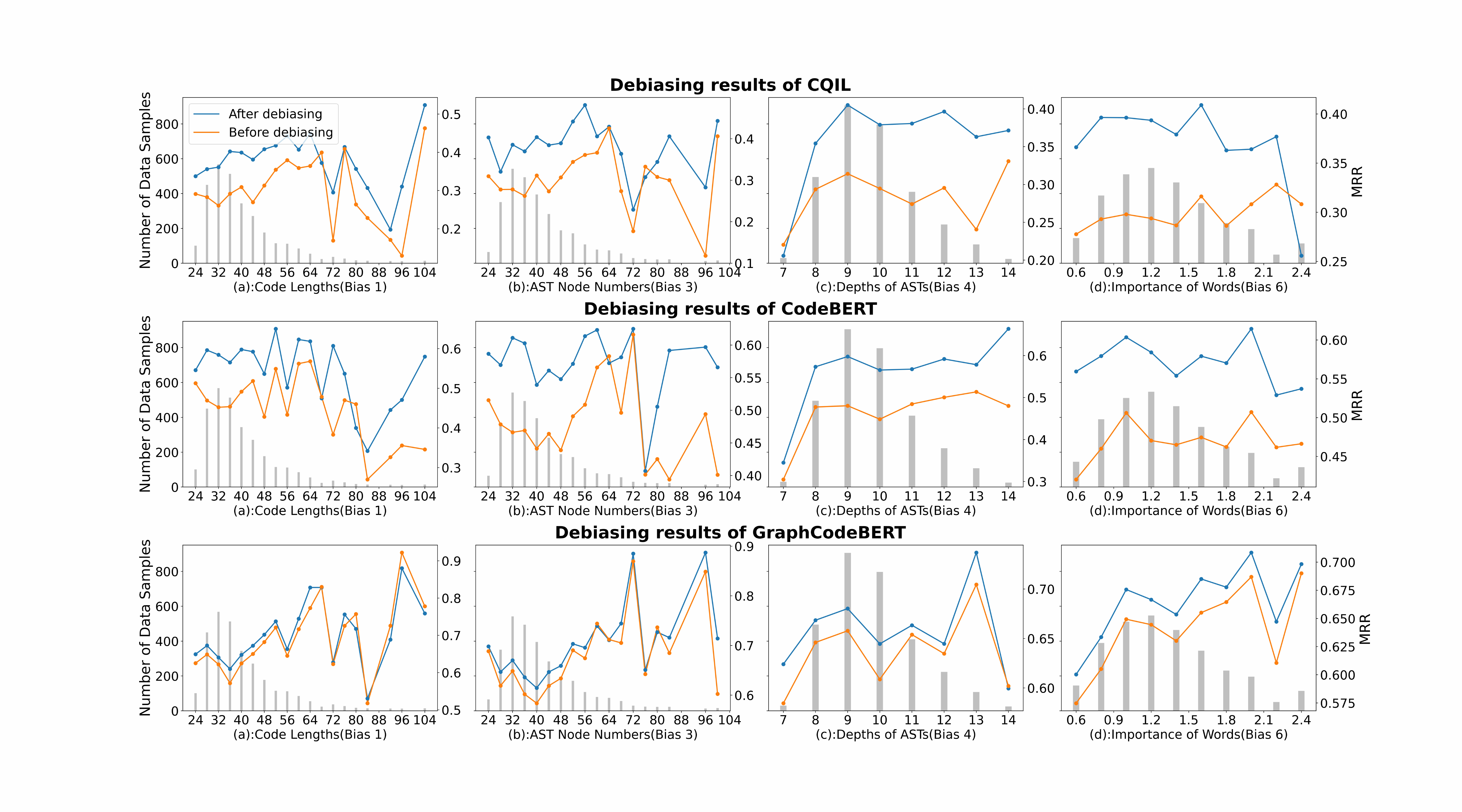}
    \vspace{-20pt}
    \caption{Mitigate biases using parallel reranking.}
    \vspace{-10pt}
    \label{fig:debiasing_parallel}
\end{figure*}

Note the above two strategies in our debiasing framework looks similar to Boosting and Bagging methods used in Ensemble Learning~\citep{Zhou09}, but they are not the same: 
(1) Compared to Boosting methods like AdaBoost~\citep{FreundS97}, sequential reranking does not increase the weights for wrongly labeled training samples (biased/unbiased cases) in previous reranker since each reranker is designed for different targets (mitigate different biases) and wrongly labeled samples in the previous reranker may be correct samples for the next reranker.
Differently, Boosting methods will increase weights of incorrectly predicted sampled for training the next learner.
(2) Compared to Bagging methods~\citep{Breiman96b}, parallel reranking does not adopt sampling to prepare different datasets (from the complete training set) for use in each reranker.
The reason is that, to make our debiasing method simple and general, our reranking method is designed as a similarity-based adjuster with simple rules instead of a learning-based approach. 
In a large training set, most similar queries that are used to judge whether current search is facing bias may not be selected in sampling, which negatively affects debiasing.


\section{Debiasing Experiment}
\label{sec:exp}

In this section, we will illustrate the effectiveness of our debiasing framework on mitigating code search biases.
Results are reported using our framework to mitigate the seven biases for the six code search methods on the CoSQA dataset.
By default, the order of rerankers in sequential reranking is Biases 7, 6, 3, 4, 2, 5 and 1.  
We also analyze the impact of reranker order in Analysis 4 of our experiments.
Our method requires three hyper-parameters: $M$, $N$ and $S$, as illustrated in Sec.~\ref{sec:single_bias}.
We search $M$, $N$ and $S$ in $\{1,3,5\}$, $\{10,15,20\}$ and $\{1,3,5\}$, respectively. Best results ($M=1$, $N=10$ and $S=1$) are reported.

\begin{table*}[!th]
    \centering
    \caption{Overall performance changes of code search models using sequential reranking.}
    \label{tab:debiasing_seq}%
    \vspace{-10pt}
    \resizebox{0.98\linewidth}{!}{
    \begin{tabular}{ccccccccc}
    \toprule
    \multirow{2}[2]{*}{Method Name} & \multicolumn{2}{c}{MRR} & \multicolumn{2}{c}{HR@1} & \multicolumn{2}{c}{HR@5} & \multicolumn{2}{c}{HR@10} \\
        & Before  & After  & Before  & After  & Before  & After  & Before  & After  \\
    \midrule
    DeepCS & 0.295 & 0.428 (+45\%) & 0.219 & 0.366 (+67\%) & 0.375 & 0.489 (+30\%) & 0.462 & 0.553 (+20\%) \\
    CQIL & 0.296 & 0.384 (+30\%) & 0.216 & 0.299 (+38\%) & 0.377 & 0.478 (+27\%) & 0.469 & 0.557 (+19\%) \\
    CodeBERT & 0.474 & 0.569 (+20\%) & 0.363 & 0.471 (+30\%) & 0.598 & 0.685 (+15\%) & 0.712 & 0.782 (+9.8\%) \\
    CoCLR & 0.756 & 0.770 (+1.9\%) & 0.641 & 0.661 (+3.1\%) & 0.909 & \ \ 0.917 (+0.88\%) & 0.967 & \ \ 0.971 (+0.41\%) \\
    GraphCodeBERT & 0.641 & 0.695 (+8.4\%) & 0.524 & 0.587 (+12\%) & 0.790 &  0.831 (+5.2\%) & 0.882 & 0.911 (+3.3\%) \\
    UniXcoder & 0.702 & 0.737 (+5.0\%) & 0.584 & 0.630 (+7.9\%) & 0.862 &  0.880 (+2.1\%) & 0.935 & \ \ 0.940 (+0.53\%) \\
    \bottomrule
    \end{tabular}%
    }
    \vspace{-5pt}
\end{table*}%

\begin{table*}[!th]
    \centering
    \caption{Overall performance changes of code search models using parallel re-ranking.}
    \label{tab:debiasing_parallel}%
    \vspace{-10pt}
    \resizebox{0.98\linewidth}{!}{
    \begin{tabular}{ccccccccc}
    \toprule
    \multirow{2}[2]{*}{Method Name} & \multicolumn{2}{c}{MRR} & \multicolumn{2}{c}{HR@1} & \multicolumn{2}{c}{HR@5} & \multicolumn{2}{c}{HR@10} \\
        & Before  & After  & Before  & After  & Before  & After  & Before  & After  \\
    \midrule
    DeepCS & 0.295 & 0.425 (+44\%) & 0.219 & 0.363 (+65\%) & 0.375 & 0.485 (+29\%) & 0.462 & 0.551 (+19\%) \\
    CQIL & 0.296 & 0.383 (+29\%) & 0.216 & 0.300 (+39\%) & 0.377 & 0.476 (+26\%) & 0.469 & 0.551 (+17\%) \\
    CodeBERT & 0.474 & 0.579 (+22\%) & 0.363 & 0.483 (+33\%) & 0.598 & 0.694 (+16\%) & 0.712 & 0.780 (+9.6\%) \\
    CoCLR & 0.756 & 0.769 (+1.7\%) & 0.641 & 0.661 (+3.1\%) & 0.909 & 0.915 (0.66\%) & 0.967 & \ \ 0.971 (+0.41\%) \\
    GraphCodeBERT & 0.641 & 0.666 (+3.9\%) & 0.524 & 0.552 (+5.3\%) & 0.790 & 0.810 (+2.5\%) & 0.882 & 0.895 (+1.5\%) \\
    UniXcoder & 0.702 & 0.716 (+2.0\%) & 0.584 & 0.602 (+3.1\%) & 0.862 & 0.872 (+1.2\%) & 0.935 & \ \ 0.939 (+0.43\%) \\
    \bottomrule
    \end{tabular}%
    }
\end{table*}%

\vspace{5pt}
\noindent\textbf{Analysis 1: Debiasing Results.}
We first analyze the results after debiasing. 
Due to space limitation, we only visualize results of Bias 1 (Lengths of Code), Bias 3 (Numbers of AST nodes), Bias 4 (Depths of ASTs) and Bias 6 (Importance of Words) for CQIL, CodeBERT and GraphCodeBERT. 
For other code search methods and biases, we observe similar results.
Fig.~\ref{fig:debiasing_seq} shows the performance before and after mitigating biases using sequential reranking.
The result using parallel reranking is presented in Fig.~\ref{fig:debiasing_parallel}.
From visualization results, we can clearly see that, for all the four biases, MRR scores of most intervals increase after deploying our debiasing framework, showing the effectiveness of our debiasing framework.
Sequential reranking shows a slightly better debiasing result than parallel reranking (e.g., see CQIL(b) and GraphCodeBERT(b) in Fig.~\ref{fig:debiasing_seq} and Fig.~\ref{fig:debiasing_parallel}).
However, sequential reranking is not as efficient as parallel reranking as it processes each reranker one by one.

\vspace{5pt}
\noindent\textbf{Analysis 2: Changes of Code Search Performance after Debiasing.}
Tab.~\ref{tab:debiasing_parallel} and Tab.~\ref{tab:debiasing_seq} illustrate the changes of overall code search performance after debiasing using sequential reranking and parallel reranking, respectively.
From results, we can see that, after debiasing, overall code search performance w.r.t. MRR or HR@K significantly increases.
The improvements are especially noticeable for DeepCS, CQIL and CodeBERT: MRR and HR@K of these methods increase by 9.6\%-67\%. The reason is that the original search performance of the three methods is not high and there is still large room for improvement.
Even for CoCLR, GraphCodeBERT and UniXcoder which show quite high MRR ($>$0.6) and HR@K ($>$0.5) before debiasing, our debiasing framework still helps improve the overall code search performance.
Thus, we can conclude that \emph{mitigating code search bias has a positive effect on improving the overall code search performance}.

\begin{figure}[t]
	\centering
	\includegraphics[width=0.95\columnwidth]{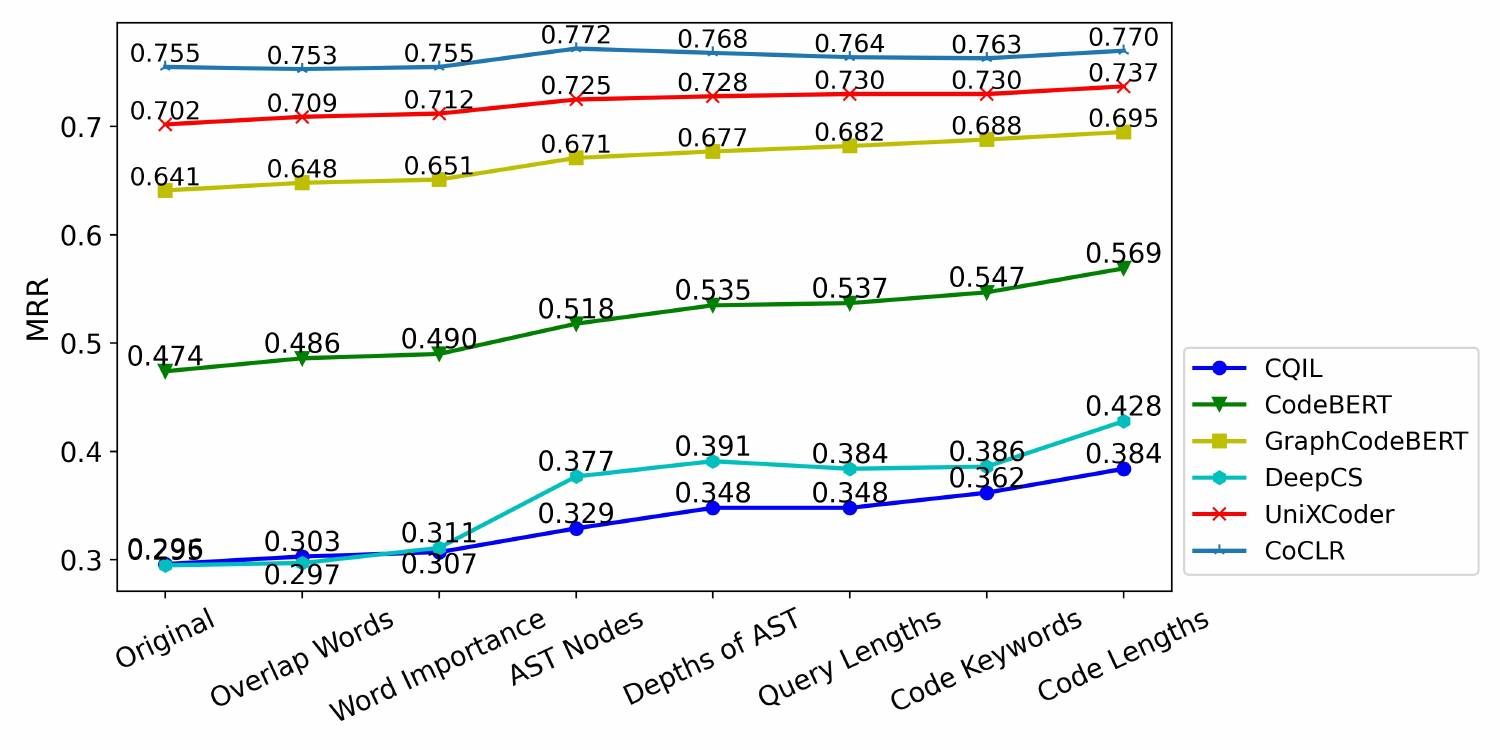}
    \vspace{-10pt}
	\caption{Changes of overall MRR after applying each reranker in sequential reranking.}
    \vspace{-10pt}
	\label{fig:MRR change}
\end{figure}

\vspace{5pt}
\noindent\textbf{Analysis 3: Impacts of Applying Multiple Rerankers.}
Next, we investigate whether applying multiple rerankers brings better debiasing results than using a single reranker. 
Fig.~\ref{fig:MRR change} illustrates the changes of overall MRR scores for the six code search models after applying each reranker using sequential reranking in the default order.
The horizontal axis labels (from left to right) show the order of rerankers applied.
We can observe that MRR scores of CodeBERT, DeepCS and CQIL gradually increase as more rerankers are applied.
Eventually, their overall performance after debiasing gets significantly improved compared to their original performance.
For CoCLR, UniXCoder and GraphCodeBERT which have achieved high MRR scores before debiasing, applying multiple rerankers slightly enhances or does not negatively affect their overall performance.
Overall, after applying seven rerankers, the performance of CoCLR, UniXCoder and GraphCodeBERT gets enhanced.
We can observe a similar trend when using parallel reranking.
In conclusion,
\emph{
the more rerankers are applied, the better overall code search performance the code search model can achieve. 
In other words, each reranker indeed contributes to the improvement of the quality of code search results.
}

\begin{figure}[t]
    \centering
    \includegraphics[width=0.8\columnwidth]{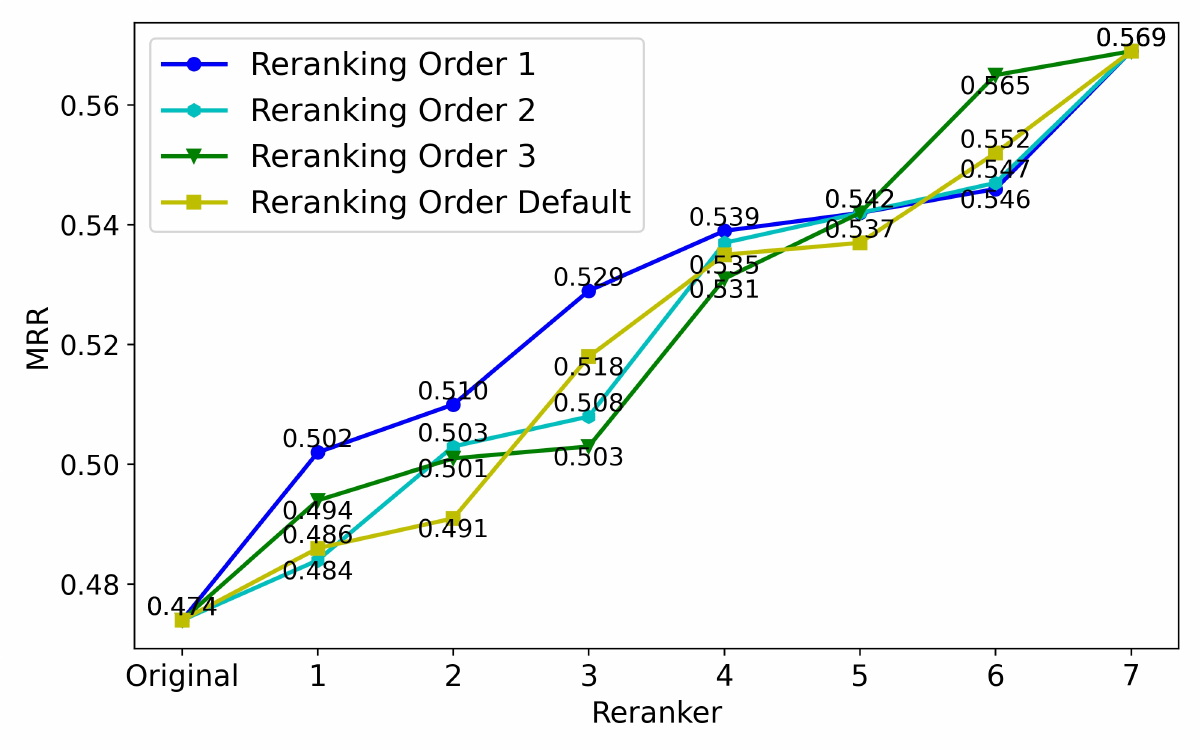}
    \vspace{-10pt}
    \caption{Sequential reranking in different order.}
    \label{fig:MRR rerank order}
    \vspace{-5pt}
\end{figure}

\vspace{5pt}
\noindent\textbf{Analysis 4: Impacts of Reranker Order in Sequential Reranking.}
Since sequential reranking has various possible order of rerankers, we analyze the impact of reranking order.
In addition to the default order, we report the debiasing performance on CodeBERT using sequential reranking with three other orders: order 1 (biases 1, 6, 4, 5, 2, 7, 3), order 2 (bases 6, 2, 4, 7, 3, 5, 1) and order 3 (biases 4, 6, 2, 1, 5, 3, 7). 
Fig.~\ref{fig:MRR rerank order} demonstrates the performance changes after each reranker is applied in the three order.
The horizontal axis labels (from left to right) show rerankers in the applied order.
Similar to the observation in Analysis 3, we can see that adding more rerankers help improve the MRR score.
And the intermediate debiasing results are slightly different using three different order.
But the different order does not affect the final debiasing result too much.

\begin{table}[t]
    \centering
    \caption{MRR for different hyper-parameters.}
    \label{tab:parameter}%
    \vspace{-10pt}
    \resizebox{\linewidth}{!}{
    \begin{tabular}{cccccccccc}
    \toprule
    \multirow{2}[2]{*}{Method} & \multicolumn{3}{c}{$M$} & \multicolumn{3}{c}{$S$} & \multicolumn{3}{c}{$N$} \\
        & 1   & 3   & 5   & 1   & 3   & 5   & 10 & 15 & 20 \\
    \midrule
    CQIL & 0.380 & 0.329 & 0.348 & 0.384 & 0.380 & 0.355 & 0.380 & 0.380 & 0.376 \\
    CodeBERT & 0.569 & 0.537 & 0.504 & 0.572 & 0.569 & 0.552 & 0.569 & 0.569 & 0.569 \\
    GraphCodeBERT & 0.695 & 0.673 & 0.660 & 0.696 & 0.695 & 0.690 & 0.695 & 0.695 & 0.695 \\
    \bottomrule
    \end{tabular}%
    }
    \vspace{-10pt}
\end{table}

\vspace{5pt}
\noindent\textbf{Analysis 5: Impacts of Hyper-Parameters.}
We further analyze the impacts of hyper-parameters.
Tab.~\ref{tab:parameter} provides the debiasing results of CQIL, CodeBERT and GraphCodeBERT using different hyper-parameters.
Each of the MRR score in the table is obtained by changing one hyper-parameter while keeping the other two hyper-parameters the same as the best ones found in hyper-parameter search.
From the result, we can conclude that hyper-parameters do not affect results too much.
We provide the analysis as follows:
\begin{itemize}[leftmargin=12pt,topsep=1pt,itemsep=0.2pt] 
\item $M$ indicates how many top-M similar queries in the training set are adopted. We believe the top-1 similar query already provides a hint for our method, and including more similar queries do not bring more information. Hence, changing $M$ does not affect the results too much.

\item $N\%$ represents the percentage of chosen training search cases with the highest MRR scores. 
Since the training set of CoSQA data contains 14K query-code pairs, changing $N\%$ in \{10$\%$, 15$\%$, 20$\%$\} results in 1,400, 2,100 and 2,800 retrieved cases, respectively. The difference between the numbers of retrieved cases is not large, compared to the total dataset with 21K query-code pairs.

\item $S$ indicates the number of clusters after performing kmeans on these $N\%$ cases.
We find that small values of $S$ bring relatively robust and good performance of debiasing, as reported in Tab.~\ref{tab:parameter}.
Therefore, we suggest that users set $S$ to a small value.
If we set $S$ to a much larger number (e.g., 100, 500, 1,000), the performance becomes inconsistent, and we suspect that dividing retrieved cases into many small clusters cannot help find case patterns. Instead, many small clusters bring the noise. Hence, we do not suggest that users set $S$ to a large value.
\end{itemize}

\vspace{5pt}
\noindent\textbf{Analysis 6: Human Evaluation.} We also conduct human evaluation for assessing the quality of debiasing. We randomly pick 200 queries from the test set for human evaluation. We choose CQIL as a representation of code search models and use it in human evaluation. We use our debiasing framework to reduce code search biases in the corresponding results of CQIL for the 200 queries. We recruit four master students majoring in computer science to check the quality of debiasing manually. For each query, we provide the students with two lists. One is the original top-10 search results from CQIL, and the other is the top-10 list after debiasing. The lists for each query are shown in random order. Students are asked to choose which top-10 list is better, and they can also indicate that the two lists are roughly of the same quality. From the results of human evaluation, we find that, for 71.5\% queries, lists after debiasing are assessed as better ones. For 19.5\% queries, the original list and the reranked list are estimated as having similar quality. For the remaining 9\% queries, debiasing degrades the quality of the search list. The human evaluation results illustrate that our debiasing method indeed improves the quality of the code search for most queries. The materials of human evaluation are included in our provided repository.


\section{Conclusion}  
In this paper, we reveal the existence of code search biases. We design a general debiasing framework that can be easily plugged into existing search models. 
In the future, we will explore pre-processing and in-processing methods to improve our framework and better mitigate code search biases.

\clearpage

\section*{Limitations}
\label{sec:lim}

This work may have some limitations:
\begin{itemize}[leftmargin=12pt,topsep=1pt,itemsep=0.2pt] 
	\item \textbf{Data:} When we submitted this manuscript, only one real code search dataset CoSQA was publicly available. Other datasets in the literature do not have real search queries, and they use code documents to simulate queries. However, code documents and queries have different text styles (i.e., length). Hence, we only study code search bias based on the real data in CoSQA. To overcome this limitation, we are constructing another dataset containing real code search queries and will release it for future study.

	\item \textbf{Language:} Queries and code snippets in CoSQA are written in English and Python, respectively. 
	It is unclear whether our analysis results hold for queries written in other natural languages (e.g., French and Chinese).
	As the causes of code search biases analyzed in this work should be common across different programming languages (e.g., Java and Go), we expect that code search in other programming languages also suffers from the biases studied in this paper. We leave the study of the impacts of different natural languages and programming languages on code search bias as future work.

\end{itemize}

\section*{Acknowledgements}
This work was partially supported by National Key R\&D Program of China (No. 2022ZD0118201), National Natural Science Foundation of China (No. 62002303, 42171456) and  CCF-Tencent Open Fund (RAGR20210129).

\bibliographystyle{acl_natbib}
\bibliography{ref.bib}

\end{document}